\documentclass[final]{cvpr}

\usepackage{times}
\usepackage{epsfig}
\usepackage{graphicx}
\usepackage{amsmath}
\usepackage{amssymb}

\usepackage{caption,subcaption}
\usepackage{multirow}
\usepackage{float}

\usepackage[pagebackref=true,breaklinks=true,colorlinks,bookmarks=false]{hyperref}

\def\cvprPaperID{26} 
\def\confYear{CVPR 2021}
\newcommand{\pastref}{\hat{\mathbf{x}}_{p}}
\newcommand{\futureref}{\hat{\mathbf{x}}_{f}}
\newcommand{\pastflow}{\mathbf{v}_{p}}
\newcommand{\futureflow}{\mathbf{v}_{f}}
\newcommand{\balph}{\boldsymbol{\alpha}}
\newcommand{\bbeta}{\boldsymbol{\beta}}
\newcommand{\pred}{\tilde{\mathbf{x}}_t}
\newcommand{\sysoutput}{\hat{\mathbf{x}}_t}
\newcommand{\code}{\mathbf{x}_t}

\begin{document}

\title{Conditional Coding and Variable Bitrate for Practical Learned Video Coding} 

\author{Th\'{e}o Ladune, Pierrick Philippe\\
Orange\\
Rennes, France\\
{\tt\small firstname.lastname@orange.com}
\and
Wassim Hamidouche, Lu Zhang, Olivier D\'{e}forges\\
CNRS, IETR -- UMR 6164\\
Univ. Rennes, INSA Rennes\\
{\tt\small firstname.lastname@insa-rennes.fr}
}
\maketitle

\begin{abstract}
    This paper introduces a practical learned video codec. Conditional coding
    and quantization gain vectors are used to provide flexibility to a single
    encoder/decoder pair, which is able to compress video sequences at a
    variable bitrate. The flexibility is leveraged at test time by choosing the
    rate and GOP structure to optimize a rate-distortion cost. Using the CLIC21
    video test conditions, the proposed approach shows performance on par with
    HEVC.   
\end{abstract}

\section{Introduction}

Deep neural networks allow to perform complex non-linear transforms. In image
coding, these transforms are learned end-to-end \cite{DBLP:conf/iclr/BalleLS17}
to minimize a rate-distortion cost. This leads to compression performance
competitive with the image coding configuration of VVC \cite{VVC_Ref}, the
latest ITU/MPEG video coding standard.
Video coding is a more challenging task than image coding due to the additional
temporal redundancies, often removed through
motion compensation. It consists in computing a temporal prediction from
already decoded frames to only send the unpredictable part. Frames
coded using already decoded ones are called \textit{inter} frames, others are
called \textit{intra} frames.\\

The unpredictable content is often conveyed through residual coding
\textit{i.e.} by sending the difference between the frame and its temporal
prediction. In the literature, this results in a \textit{two-codec} approach,
with an inter-frame codec dedicated to residual signal and an intra-frame codec
for image-domain signal
\cite{Agustsson_2020_CVPR,yang2020learning,yilmaz2020endtoend}.
Moreover, previous learned video coders
\cite{Agustsson_2020_CVPR,DBLP:conf/cvpr/LuO0ZCG19,yang2020learning}
are designed to operate under a single rate constraint (\textit{i.e.} a single
quality level). As such, having several quality levels available requires the
storage of one decoder per quality level, which is not practical for real world
application. \\

This article introduces an end-to-end learned factorized system. It is claimed
to be a practical video coder as it provides some essential real-world features.
First, a single coder processes intra and inter frames, allowing to use any GOP
structures (intra/inter arrangement). Second, a single coder allows to
continuously select the rate of each frame, enabling the accurate optimization
of each frame RD cost.\\

The coding scheme is based on two networks, MOFNet and CodecNet
\cite{ladune2021conditional}. MOFNet conveys side-information (prediction parameters,
coding mode) while CodecNet transmits the unpredictable content. Both networks
use conditional coding to exploit the decoder-side information while being
resilient to its absence. This enables the processing of inter and intra frames
with the same coder. Gain vectors \cite{variablerate} are added at the
quantization step of both networks to adapt to different rate constraints. The
system is shown to be competitive with the video coding standard HEVC
\cite{Sullivan:2012:OHE:2709080.2709221} under the CLIC21 video test conditions
\cite{CLIC21}.

\section{Coding Scheme}

\begin{figure}[H]
    \includegraphics[width=\linewidth]{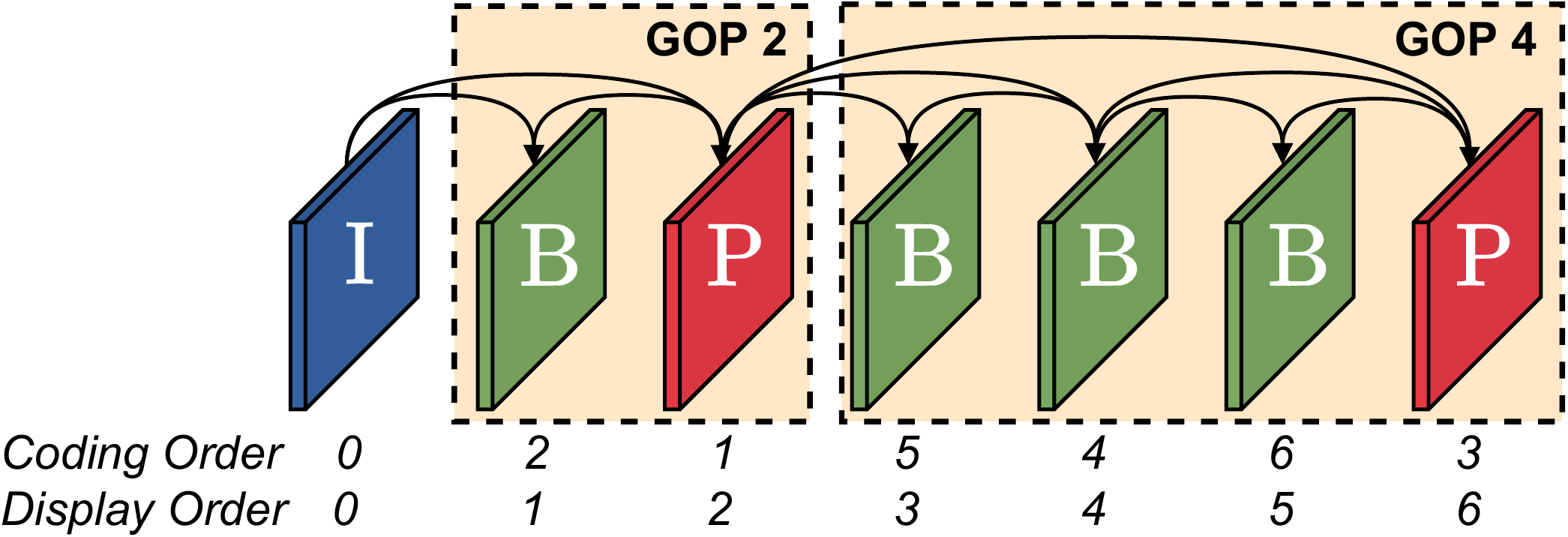}
    \caption{Two different GOP structures, GOP2 and GOP4}
    \label{fig:coding_config}
\end{figure}

Let $\left\{\mathbf{x}_i \in \mathbb{R}^{C \times H \times W}, i \in
\mathbb{N}\right\}$ be a video with $C$ color channels of height $H$
and width $W$\footnote{Videos are in YUV 420. For convenience, a bilinear
upsampling is used to obtain YUV 444 data.}. In this work, videos are
encoded using one intra frame (I-frame) followed by several Groups of Pictures
(GOP). A GOP is a fixed pattern of inter-frames, made of P-frames  (use one
already decoded frame as reference) and B-frames (use two references). The Fig.
\ref{fig:coding_config} shows an I-frame and two different GOP
structures.\\
\vfill
\vfill

\begin{figure*}[ht]
    \centering
    \includegraphics[width=\linewidth]{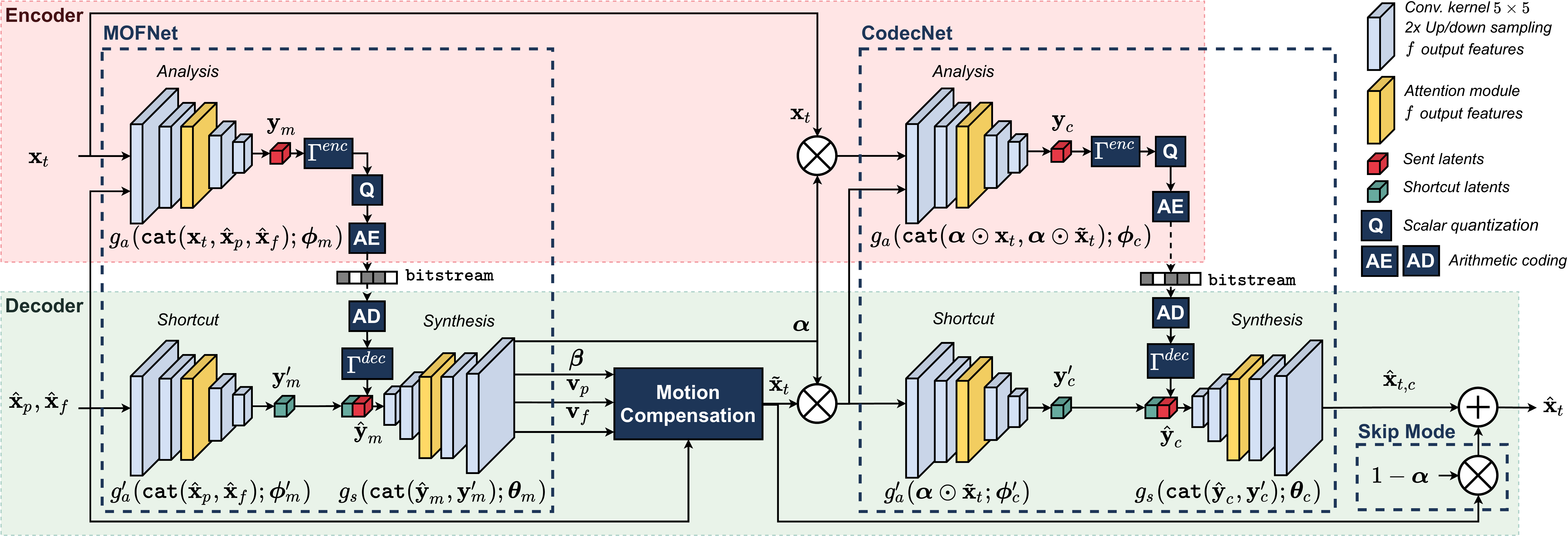}
    \caption{Diagram of the system. Latents probability model uses hyperpriors
    \cite{DBLP:conf/iclr/BalleMSHJ18} and is omitted for clarity. Attention modules are implemented as proposed
    in \cite{cheng2020learned} and $f = 192$. There are 28 millions learnable
    parameters $\left\{\boldsymbol{\phi},\boldsymbol{\theta}\right\}$.}
    \label{fig:overalldiagram}
\end{figure*}

\newpage
Video sequences are compressed to minimize a rate-distortion trade-off weighted
by $\lambda$:
\begin{equation}
    \mathcal{L}_\lambda = \sum_t \mathrm{D}(\sysoutput, \code) + \lambda \mathrm{R}(\sysoutput),
    \label{eq:rd}
\end{equation}
with $\mathrm{D}$ a distortion measure between the original frame $\code$ and
the coded one $\sysoutput$ and $\mathrm{R}(\sysoutput)$ the rate associated to
$\sysoutput$. Following the CLIC21 video test conditions,  $\mathrm{D}$ is
based on the MS-SSIM \cite{Wang03multi-scalestructural}:
$\mathrm{D}(\sysoutput, \code) = 1 - \text{MS-SSIM}(\sysoutput, \code)$.

\subsection{Overview}

Let $\code$ be a frame to code. To reduce its rate, the system leverages up to
two reference frames denoted as $(\pastref, \futureref)$. The proposed coding
scheme, shown in Fig. \ref{fig:overalldiagram}, is split between two
convolutional networks, MOFNet and CodecNet. MOFNet takes $(\code, \pastref,
\futureref)$ as inputs to compute and convey two optical flows $(\pastflow,
\futureflow)$, a pixel-wise prediction weighting $\bbeta$ and a pixel-wise
coding mode selection $\alpha$. Each optical flow represents a pixel-wise motion
from $\code$ to one of the reference, used to interpolate the reference using a
bilinear warping. Both reference warpings are summed and weighted by $\bbeta$ to
obtain a temporal prediction $\pred$:
\begin{equation}
    \pred = \bbeta \odot w(\pastref;\pastflow) + (1 - \bbeta) \odot w(\futureref;\futureflow),
\end{equation}
where $w$ is a bilinear warping, $\odot$ a pixel-wise multiplication, $\bbeta \in \left[0, 1\right]^{H \times
W}$ and $\pastflow, \futureflow \in \mathbb{R}^{2\times H \times W}$.\\

The coding mode selection $\balph \in \left[0, 1\right]^{H \times
W}$ arbitrates between two coding modes: \textit{Skip} mode (direct copy of
$\pred$) or CodecNet to transmit $\code$ using information from $\pred$ to reduce
the rate. The contributions from the two coding modes are summed to obtain the
reconstructed frame $\sysoutput$:
\begin{equation}
    \sysoutput = \underbrace{(1 - \balph) \odot \pred}_{\text{Skip}} +
    \underbrace{c(\balph \odot \mathbf{x}_t, \balph \odot \pred)}_{\text{CodecNet}}.
\end{equation}
The total rate is the sum of MOFNet and CodecNet rate. When there is no
reference available, MOFNet is bypassed, $\pred$ is set to 0 and $\balph$ to 1:
image coding is de facto used.

\subsection{Conditional Coding}

\newcommand{\rootpathgain}{/home/theo/PhD/Article/CVPR21/paper/images/tikz/gain_vector}
\begin{figure*}[htb]
    \centering
    \begin{subfigure}{0.49\textwidth}
        \centering
        \includegraphics[width=0.85\linewidth]{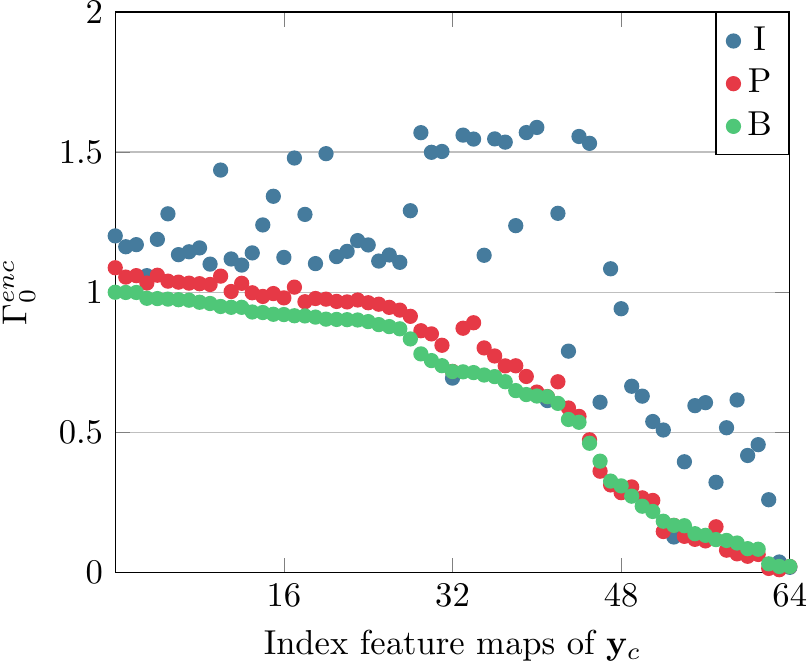}
        \caption{$\Gamma_0^{enc}$ learned under the rate constraint $\lambda_0$ (higher rate).}
        \label{fig:gain_vector_plot_0}  
    \end{subfigure}
    \begin{subfigure}{0.49\textwidth}
        \centering
        \includegraphics[width=0.85\linewidth]{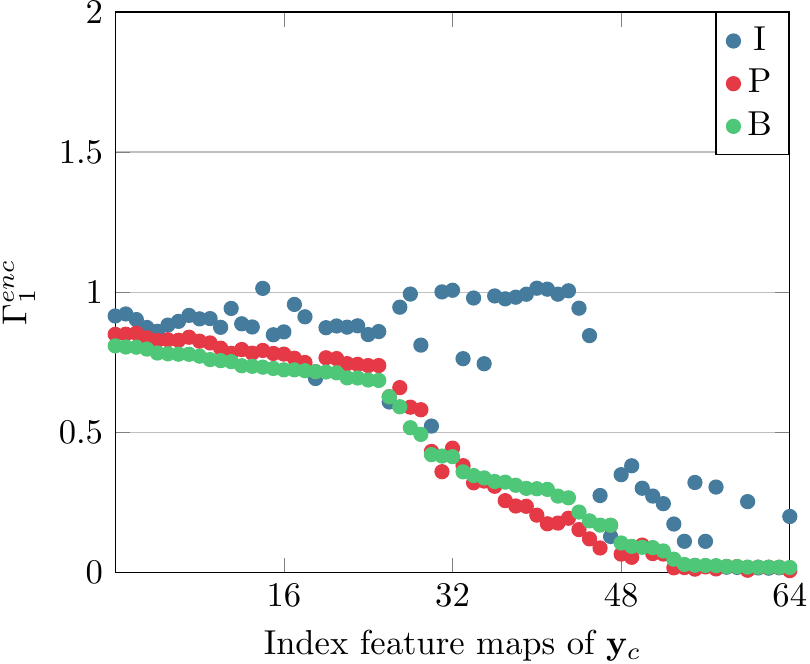}
        \caption{$\Gamma_1^{enc}$ learned under the rate constraint $\lambda_1$ (lower rate).}
        \label{fig:gain_vector_plot_1}  
    \end{subfigure}
    \caption{Feature-wise value of the CodecNet gain vectors $\Gamma_0^{enc}$
    and $\Gamma_1^{enc}$, corresponding to two rate constraints $\lambda_0$
    (higher rate) and $\lambda_1$ (lower rate). The graphs present the gain vectors
    associated to I, P and B-frames.}
    \label{fig:gain_vector_plot}
\end{figure*}

MOFNet and CodecNet rely on conditional coding \cite{ladune2021conditional} to better
exploit decoder-side information than residual coding. Conditional coding
supplements the analysis and synthesis transforms of the auto-encoder
architecture \cite{DBLP:conf/iclr/BalleMSHJ18} with a \textit{shortcut}
transform. The shortcut transform retrieves information from the references
(\textit{i.e.} at no rate) as latents $\mathbf{y}^\prime$. Thus, the analysis
transform sends only the information not present at the decoder as latents
$\mathbf{y}$. Finally, the synthesis transform concatenates both latents as
inputs to compute its output. After training, conditional coding is flexible
enough to work with zero, one or two references. This makes the proposed coder
able to process all types of frames, allowing to use any GOP structure composed
of I, P and B-frames. 

\subsection{Variable rate}

\begin{figure}[H]
    \centering
    \includegraphics[width=0.68\linewidth]{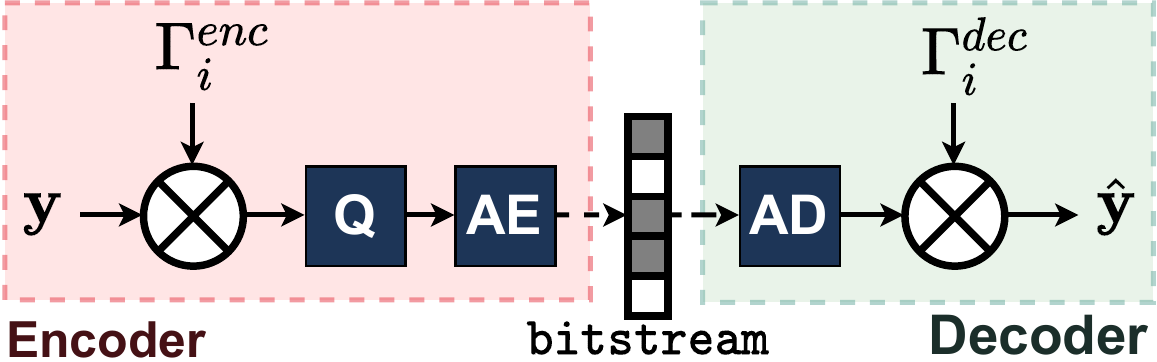}
    \caption{Usage of the quantization gain vectors.}
    \label{fig:gain_vector}
\end{figure}

The system operates at variable rates using scaling operations in the
quantization stage \cite{variablerate}. The coder is trained simultaneously for
$N$ rate constraints $\left\{\lambda_1,\ldots,\lambda_N\right\}$. Each
$\lambda_i$ is associated to a pair of learned feature-wise gain vectors
$\Gamma_i^{enc}, \Gamma_i^{dec} \in \mathbb{R}^{C_y}$, with $C_y$ the number of
channels of the latents $\hat{\mathbf{y}}$. The gain vectors are applied through
feature-wise multiplication (see Fig. \ref{fig:gain_vector}).\\

The gain vectors can be thought of as feature-wise adapted quantization
steps. Since the quantization steps of I, P and B frames may be different, a
dedicated gain vector pair is learned for each frame type and for each rate
constraint, for both MOFNet and CodecNet. After training, it is possible to
operate under any \textit{continuous} rate constraint $r \in \left[1, N\right]$
by interpolating the gain vectors through a weighted geometric averaging:
\begin{equation}
    \Gamma_r = \Gamma_{\lfloor r\rfloor}^{\ 1 - l}\odot\
    \Gamma_{\lceil r\rceil}^{\ l}, \text{ with } l = r - \lfloor r\rfloor.
    \label{eq:gain_interpol}
\end{equation}

\section{Training}

The purpose of the training phase is to prepare the system to code I, P and
B-frames under $N = 6$ rate constraints. To this effect, it is trained
on the smallest coding structure featuring the 3 types of frames \textit{i.e.} an
I-frame followed by a GOP2 (see Fig. \ref{fig:coding_config}). For each training
iteration, a rate index $i \in \left\{1, N \right\}$ is randomly selected. Then the
3 frames are coded using the corresponding $\Gamma_i$, followed by a single
back-propagation to minimize the rate-distortion cost of eq. \eqref{eq:rd} using
$\lambda_i$. The $N$ rate constraints are chosen to be distributed
around the 1 Mbit/s rate target of CLIC21.\\

 The learning process is performed through
a rate-distortion loss. No element of the system requires a pre-training or a
dedicated loss. Moreover, coding the 3 frames in the forward pass makes the
system able to deal with coded references, leading to better coding
performance.

\section{Experimental Results}
\newcommand{\rootpathcodingconfigcompete}{/home/theo/PhD/Article/CVPR21/paper/images/tikz/coding_struct_compete}
\begin{figure*}[htb]
    \centering
    \begin{subfigure}{0.45\textwidth}
        \centering
        \includegraphics[width=0.9\linewidth]{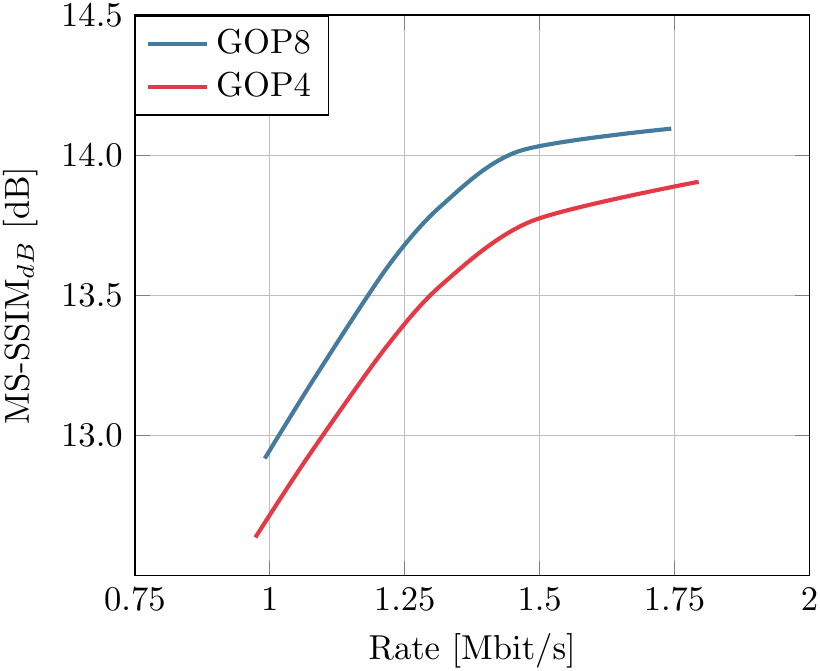}
        \caption{\textit{VR\_2160P-5489}. Further reference frames are beneficial.}
        \label{fig:rd_seq_A}  
    \end{subfigure}
    \begin{subfigure}{0.45\textwidth}
        \centering
        \includegraphics[width=0.9\linewidth]{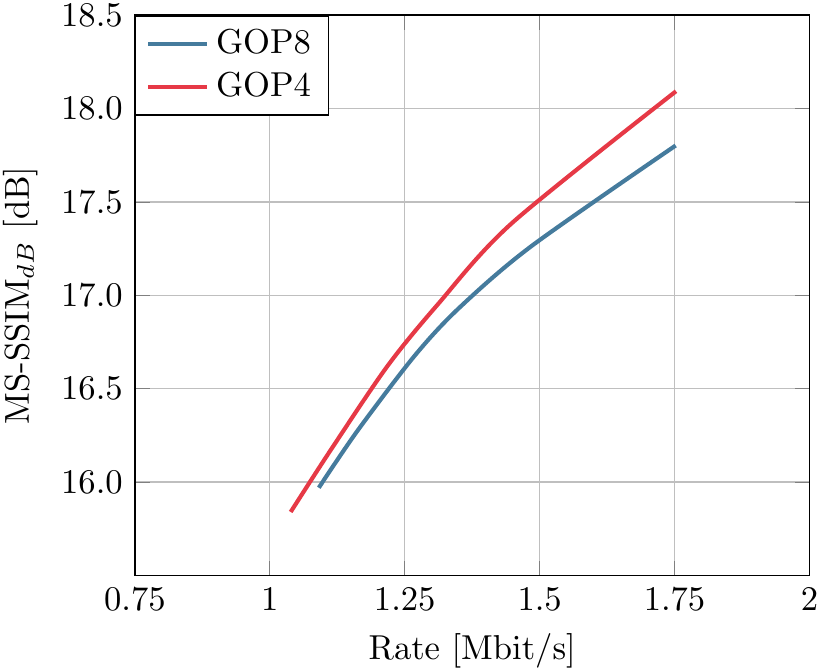}
        \caption{\textit{VR\_1080P-384e}. Further reference frames are harmful.}
        \label{fig:rd_seq_B}  
    \end{subfigure}
    \caption{Rate-distortion curves on two CLIC21 sequences for two different
    GOP structures. The intra period is set to 32 frames for both GOP
    structures. The quality is measured as MS-SSIM$_{dB} = -10\log_{10}(1 -
    $MS-SSIM$)$, higher is better.}
    \label{fig:rd_coding_struct}
\end{figure*}

\subsection{Gain Vector Visualization}

The encoder gain vectors of CodecNet obtained after training are illustrated in
Figures \ref{fig:gain_vector_plot_0} (higher rate) and
\ref{fig:gain_vector_plot_1} (lower rate). $\Gamma^{enc}$ for the I-frames are
always larger than those of the P-frames which are almost identical to those of
the B-frames. As a bigger $\Gamma^{enc}$ translates to a more accurate
quantization of $\mathbf{y}$, it means that I-frames require more precision
(\textit{i.e.} more rate) than the P or B-frames. This is because I-frames are
exclusively conveyed by CodecNet latents $\mathbf{y}_c$ whereas P and B-frames
rely on Skip Mode and CodecNet shortcut transform, allowing them to use a less
accurate quantization step. Comparing $\Gamma^{enc}_0$ and $\Gamma^{enc}_1$
illustrates that rate control is achieved through changes in the quantization
precision. As expected, a less accurate quantization is used at lower rate.


\subsection{Rate-distortion Results}
The rate and GOP structure flexibility provided by the coder allows to tune
these two parameters to optimize the rate-distortion (RD) cost individually for
each video. The Fig. \ref{fig:rd_coding_struct} shows the RD curves of the coder
on two sequences from the CLIC21 validations set for GOP4 and GOP8 structures.
This example shows that there is no consistent best GOP structure. Thanks to its
flexibility, the proposed coder can test different GOP structures during the
encoding process and select the best for each sequence, resulting in better
overall performances. Gain vectors interpolation presented in eq.
\eqref{eq:gain_interpol} makes the coder able to target any rate, allowing to
obtain continuous RD curves.\\

\begin{figure}[htb]
    \includegraphics[width=0.9\linewidth]{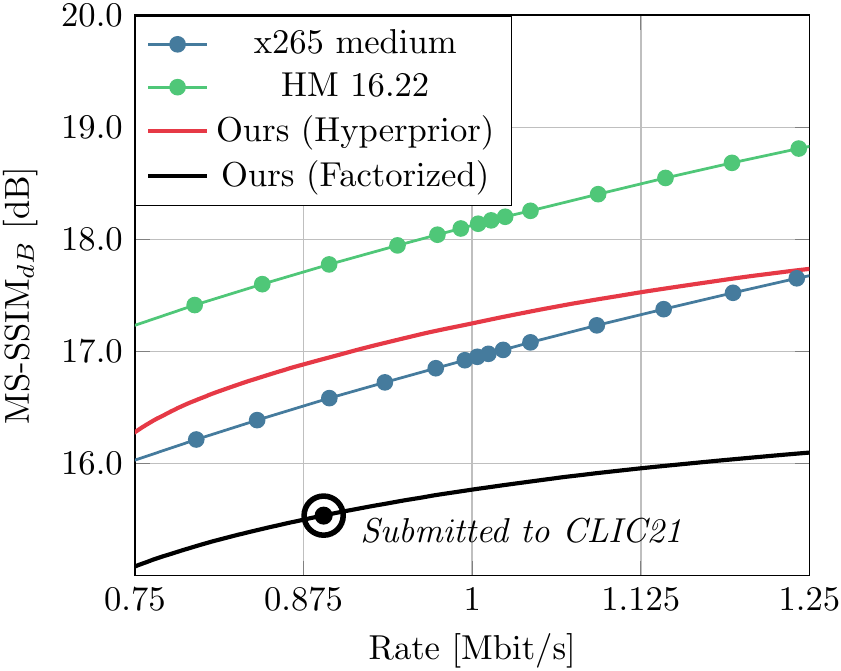}
    \caption{Rate-distortion curves on the CLIC21 video validation set. Two
    versions of the system are compared against two implementations of HEVC. The
    quality is measured as MS-SSIM$_{dB} = -10\log_{10}(1 - $MS-SSIM$)$, higher
    is better.}
    \label{fig:rd_results}
\end{figure}

The proposed system is evaluated under the CLIC21 video test conditions. The
objective is to get the highest MS-SSIM at about 1 Mbit/s. The CLIC21 validation
set contains 100 videos with 60 frames. The coder flexibility is leveraged by
optimizing the rate and the GOP structure of each sequence according to a global
RD cost. The \textit{Hyperprior} curve in Fig. \ref{fig:rd_results} shows the
performance of the coder against two implementations of the video coding
standard HEVC: the widely used \texttt{x265}\footnote{\texttt{ffmpeg
-video\_size WxH -i in.yuv -c:v libx265 -pix\_fmt yuv420p -crf QP -preset medium
-tune ssim out.mp4}} and the HM 16.22, the reference HEVC coder. The proposed
system offers compelling compression results, consistently outperforming
\texttt{x265} from 0.75 to 1.25 Mbit/s.

\vfill
\vfill

\section{CLIC21 Video Track}

A system alike to the one described in this article has been submitted to the
CLIC21 video track under the name E2E\_T\_OL. Because of the model size penalty,
the number of convolution features $f$ is reduced from 192 to 128. In order to
ensure cross-platform arithmetic coding, the submitted coder rely on a simpler
latents probability model, replacing the hyperprior mechanism with a factorized
model \cite{DBLP:conf/iclr/BalleMSHJ18}. The degradation of compression
efficiency associated to the factorized model is shown in Fig
\ref{fig:rd_results}.

\begin{table}[H]
    \centering
    \caption{CLIC21 leaderboard results validation set}
    \begin{tabular}{c||c|cc|c}
        System      & \multirow{2}{*}{MS-SSIM} &  \multicolumn{2}{c|}{Size [KBytes]} & Decoding \\
        name        &         & Data   & Model   & time [s] \\
        \hline   
        E2E\_T\_OL  & 0.97204 & 23769  & 45235   & 13356         \\
    \end{tabular}
\label{table:clic_leaderboard}
\end{table}

\section{Conclusion}

This paper introduces a end-to-end learned video coder, integrating many features
required for a practical usage. Thanks to conditional coding and quantization gain
vectors, a single encoder/decoder pair is able to process all types of frame (I,
P \& B) at any continuous rate target. Aside from being more convenient, this
flexibility brings performance gains through rate and GOP structure
competition. The relevance of the proposed approach is proved by outperforming
\texttt{x265} on the CLIC21 video coding track.\\

Immediate future work will consist in the quantization of the network operations
in order allowing the usage of hyperpriors in a cross-platform situation.

{\small
\bibliographystyle{ieee_fullname}
\bibliography{refs}

\begin{thebibliography}{10}\itemsep=-1pt

\bibitem{Agustsson_2020_CVPR}
Eirikur Agustsson, David Minnen, Nick Johnston, Johannes Balle, Sung~Jin Hwang,
  and George Toderici.
\newblock Scale-space flow for end-to-end optimized video compression.
\newblock In {\em Proceedings of the IEEE/CVF Conference on Computer Vision and
  Pattern Recognition (CVPR)}, June 2020.

\bibitem{DBLP:conf/iclr/BalleLS17}
Johannes Ball{\'{e}}, Valero Laparra, and Eero~P. Simoncelli.
\newblock End-to-end optimized image compression.
\newblock In {\em 5th International Conference on Learning Representations,
  {ICLR} 2017, Toulon, France}, 2017.

\bibitem{DBLP:conf/iclr/BalleMSHJ18}
Johannes Ball{\'{e}}, David Minnen, Saurabh Singh, Sung~Jin Hwang, and Nick
  Johnston.
\newblock Variational image compression with a scale hyperprior.
\newblock In {\em 6th International Conference on Learning Representations,
  {ICLR} 2018, Vancouver, BC, Canada}, 2018.

\bibitem{cheng2020learned}
Zhengxue Cheng, Heming Sun, Masaru Takeuchi, and Jiro Katto.
\newblock Learned image compression with discretized gaussian mixture
  likelihoods and attention modules, 2020.

\bibitem{variablerate}
T. {Guo}, J. {Wang}, Z. {Cui}, Y. {Feng}, Y. {Ge}, and B. {Bai}.
\newblock Variable rate image compression with content adaptive optimization.
\newblock In {\em 2020 IEEE/CVF Conference on Computer Vision and Pattern
  Recognition Workshops (CVPRW)}, pages 533--537, 2020.

\bibitem{VVC_Ref}
S.~Kim J.~Chen, Y.~Ye.
\newblock Algorithm description for versatile video coding and test model 8
  (vtm 8), Jan. 2020.

\bibitem{ladune2021conditional}
Th{\'e}o Ladune, Pierrick Philippe, Wassim Hamidouche, Lu Zhang, and Olivier
  D{\'e}forges.
\newblock Conditional coding for flexible learned video compression.
\newblock In {\em Neural Compression: From Information Theory to Applications
  -- Workshop @ ICLR 2021}, 2021.

\bibitem{DBLP:conf/cvpr/LuO0ZCG19}
Guo Lu, Wanli Ouyang, Dong Xu, Xiaoyun Zhang, Chunlei Cai, and Zhiyong Gao.
\newblock {DVC:} an end-to-end deep video compression framework.
\newblock In {\em {IEEE} Conference on Computer Vision and Pattern Recognition,
  {CVPR} 2019, Long Beach, CA, USA, 2019}, pages 11006--11015, 2019.

\bibitem{Sullivan:2012:OHE:2709080.2709221}
Gary~J. Sullivan, Jens-Rainer Ohm, Woo-Jin Han, and Thomas Wiegand.
\newblock Overview of the high efficiency video coding (hevc) standard.
\newblock {\em IEEE Trans. Cir. and Sys. for Video Technol.},
  22(12):1649--1668, Dec. 2012.

\bibitem{Wang03multi-scalestructural}
Zhou Wang, Eero~P. Simoncelli, and Alan~C. Bovik.
\newblock Multi-scale structural similarity for image quality assessment.
\newblock In {\em in Proc. IEEE Conf. on Signals, Systems, and Computers},
  pages 1398--1402, 2003.

\bibitem{CLIC21}
Workshop and Challenge on Learned Image~Compression.
\newblock https://www.compression.cc/, June 2021.

\bibitem{yang2020learning}
Ren Yang, Fabian Mentzer, Luc~Van Gool, and Radu Timofte.
\newblock Learning for video compression with hierarchical quality and
  recurrent enhancement, 2020.

\bibitem{yilmaz2020endtoend}
M.~Akin Yilmaz and A.~Murat Tekalp.
\newblock End-to-end rate-distortion optimization for bi-directional learned
  video compression, 2020.

\end{thebibliography}
}

\end{document}